\documentclass[conference]{IEEEtran}

\IEEEoverridecommandlockouts
\usepackage{cite}
\usepackage{amsmath,amssymb,amsfonts}
\usepackage{algorithmic}
\usepackage{graphicx}
\usepackage{textcomp}
\usepackage{xcolor}
\def\BibTeX{{\rm B\kern-.05em{\sc i\kern-.025em b}\kern-.08em
    T\kern-.1667em\lower.7ex\hbox{E}\kern-.125emX}}

\makeatletter
\DeclareRobustCommand\onedot{\futurelet\@let@token\@onedot}
\def\@onedot{\ifx\@let@token.\else.\null\fi\xspace}
\usepackage{xspace}
\newif\ifwithlatentalgo
\newif\iflong
\newif\iflettrine
\newif\ifiksvm
\newif\ifaptable
\longtrue
\newif\ifmultitables
\multitablestrue
\newif\ifappendix
\appendixtrue
\newif\iflongtables
\newif\ifltpap
\newif\ifnooverview

\def\lone{L1\xspace}
\def\smoothl{SmoothL1\xspace}
\def\logl{LogL1\xspace}
\def\cosl{Cosine\xspace}
\def\kld{KLD\xspace}
\def\spw{$ \ell_{spw}$\xspace}
\def\stpw{$ \ell_{stpw} $\xspace}
\def\fe{$ \ell_{fe}$\xspace}
\def\cv{$ \ell_{cv}$\xspace}
\def\ca{$ \ell_{ca}$\xspace}

\def\dnn{deep neural networks\xspace}

\def\psm{PSMNet\xspace}

\def\acvnet{ACVNet\xspace}
\def\gwcnet{GwcNet\xspace}
\def\leastereo{LEAStereo\xspace}
\def\cfnet{CFNet\xspace}
\def\lacnet{Lac-Net\xspace}

\def\kd{knowledge distillation\xspace}
\def\Kd{Knowledge distillation\xspace}
\def\KD{Knowledge Distillation\xspace}

\def\macs{MACs\xspace}

\def\kitti{KITTI\xspace}
\def\kittif{KITTI\xspace2015\xspace}

\def\sf{SceneFlow\xspace}
\def\ethz{ETH3D\xspace}
\def\mb{Middlebury\xspace}

\def\sota{state-of-the-art\xspace}

\def\twd{2\Db\xspace}
\def\td{3\Db\xspace}
\def\fd{4\Db\xspace}

\DeclareMathAlphabet{\pazocal}{OMS}{zplm}{m}{n}

\newcommand{\Db}{\ensuremath{\pazocal{D}}}

\def\tabref#1{Table~\ref{table:#1}}
\def\figref#1{Fig.~\ref{fig:#1}}

\def\opix{1-\textit{px}\xspace}
\def\tpix{2-\textit{px}\xspace}
\def\dpix{3-\textit{px}\xspace}
\def\fpix{4-\textit{px}\xspace}
\def\dsnet{DSNet\xspace}
\def\dsnetpa{DSNet+Attention\xspace}

\def\bbo{BB$_{21}$\xspace}
\def\bbt{BB$_{18}$\xspace}
\def\bbd{BB$_{14}$\xspace}
\def\bmark{BB$_{56}$\xspace}
\def\epe{EPE\xspace}

\makeatletter
\DeclareRobustCommand\onedot{\futurelet\@let@token\@onedot}
\def\@onedot{\ifx\@let@token.\else.\null\fi\xspace}

\def\ie{\emph{i.e}\onedot}
\def\Ie{\emph{I.e}\onedot}

\def\wrt{\emph{w.r.t}\onedot}

\def\cf{\emph{c.f}\onedot}

\def\etc{\emph{etc}\onedot}

\def\eg{\emph{e.g}\onedot}

\def\para#1{{\bf #1.}}

\def\mbold#1{\textbf{#1}}
\def\mitalic#1{\textit{#1}}

\newif\ifwithcomments 
\withcommentsfalse
\ifwithcomments

\long\def\comment#1{\para{\sffamily\{***\color{RoyalBlue}#1}****\} }%

\long\def\discuss#1{{\color{ForestGreen}\mbold{\underline{DISCUSS:}}#1} }
\long\def\diagrams#1{\\{\color{Red}\mbold{\underline{Diagram:}}#1} }

\long\def\idea#1{{\\\color{Magenta} \mbold{Idea:}#1}}
\long\def\revcomment#1{Reviewer BMVC:#1 }
\else

\long\def\comment#1{}
\long\def\discuss#1{ }
\long\def\revcomment#1{ }
\long\def\diagrams#1{ }
\long\def\idea#1{}
\fi

\def\l2norm{\mitalic{L2Norm}}

\def\figref#1{Fig.~\ref{fig:#1}}

\def\tabref#1{Table~\ref{table:#1}}
\def\secref#1{Sec.~\ref{sec:#1}}

\def\eqref#1{Eq.~(\ref{eq:#1})}

\def\ms{\textit{ms}\xspace}

\newcommand{\ltwo}{\ensuremath{L_{2}}\xspace}

\newcounter{rno}

\definecolor{darkred}{rgb}{0.8,0,0}
\definecolor{darkgreen}{rgb}{0,0.5,0}
\definecolor{darkblue}{rgb}{0,0,0.7}
\definecolor{darkpurple}{rgb}{0.4,0,0.6}
\definecolor{lightgray}{rgb}{0.92,0.92,0.92}
\definecolor{lightpink}{rgb}{1.00,0.90,0.90}

\definecolor{ellisred}{rgb}{0.87,0.44,0.38}
\definecolor{ellisgreen}{rgb}{0.69,0.90,0.52}
\definecolor{elliscyan}{rgb}{0.29,0.77,0.74}
\definecolor{ellisorange}{rgb}{0.89,0.55,0.28}
\definecolor{ellisblue}{rgb}{0.41,0.61,0.86}

\usepackage{times}
\usepackage{epsfig}
\usepackage{graphicx}
\usepackage{amsmath}
\usepackage{amssymb}
\usepackage{booktabs}
\usepackage{subcaption}

\def\tabref#1{Table~\ref{table:#1}}
\def\figref#1{Fig.~\ref{fig:#1}}

\def\opix{$ 1 $\textit{px}\xspace}
\def\tpix{$ 2 $\textit{px}\xspace}
\def\dpix{$ 3 $\textit{px}\xspace}
\usepackage{siunitx}
\usepackage{numprint}
\usepackage{makecell}
\usepackage{multirow}
\usepackage{arydshln, booktabs}
\usepackage{siunitx}
\usepackage{subcaption}
\usepackage{enumitem}
\usepackage{rotating}
\usepackage{adjustbox}
\usepackage{xcolor}
\usepackage{soul}
\usepackage{xcolor}
\usepackage{color}
\usepackage{colortbl}
\usepackage{tablefootnote}
\usepackage[absolute,overlay]{textpos}

\begin{document}

\title{Distilling Stereo Networks for Performant and Efficient Leaner Networks\\
\thanks{Funded by the Deutsche Forschungsgemeinschaft (DFG, German Research Foundation) under Germany’s Excellence Strategy – EXC number 2064/1 – Project number 390727645}}

\author{\IEEEauthorblockN{Rafia Rahim}
\IEEEauthorblockA{\textit{Cognitive Systems Group} \\
	\textit{University of Tuebingen}\\
	Tuebingen, Germany \\
	rafia.rahim@uni-tuebingen.de}
\and
\IEEEauthorblockN{Samuel Woerz}
\IEEEauthorblockA{
	\textit{University of Tuebingen}\\
	Tuebingen, Germany \\
	samuel.woerz@student.uni-tuebingen.de}
\and
\IEEEauthorblockN{Andreas Zell}
\IEEEauthorblockA{\textit{Cognitive Systems Group} \\
	\textit{University of Tuebingen}\\
	Tuebingen, Germany \\
	andreas.zell@uni-tuebingen.de}
}
\begin{textblock*}{21cm}(0.2cm, 26.7cm) 
	\noindent
	\copyright \ 2023 IEEE. Personal use of this material is permitted. Permission from IEEE must be obtained for all other uses, in any current or future media, including reprinting/republishing this material for advertising or promotional purposes, creating new collective works, for resale or redistribution to servers or lists, or reuse of any copyrighted component of this work in other works.
\end{textblock*}

\maketitle

\begin{abstract}
\Kd has been quite popular  in vision for tasks like classification and segmentation however not much work has been done for distilling \sota stereo matching methods despite their range of applications. One of the reasons for its lack of use in stereo matching networks is due to the inherent complexity of these networks, where a typical network is composed of multiple two- and three-dimensional modules. In this work, we systematically combine the insights from \sota stereo methods with general knowledge-distillation techniques to develop a joint framework for stereo networks distillation with competitive results and faster inference. Moreover, we show, via a detailed empirical analysis, that distilling knowledge from the stereo network requires careful design of the complete distillation pipeline starting from backbone to the right selection of distillation points and corresponding loss functions.  This results in the student networks that are not only leaner and faster but give excellent performance . For instance, our student network while performing better than the performance oriented methods like \psm\cite{chang_pyramid_2018}, \cfnet \cite{shen2021cfnet}, and \leastereo~\cite{cheng2020hierarchical}) on benchmark \sf dataset, is $ 8\times $, $ 5\times $, and $8\times $ faster respectively. 
Furthermore, compared to speed oriented methods having inference time less than 100\ms, our student networks perform better than all the tested methods. In addition, our student network also shows better generalization capabilities when tested on unseen datasets like \ethz and \mb\footnote{Code: https://github.com/cogsys-tuebingen/Distilling-Stereo-Networks}.
\end{abstract}

\begin{IEEEkeywords}
Knowledge Distillation, Stereo Matching, Depth Estimation
\end{IEEEkeywords}


\section{Introduction}
%


Estimating depth using a pair of stereo images is a popular depth estimation method mainly because it is cheaper compared to expensive sensors like LiDARs \cite{poggi_synergies_2020} and is more accurate than the single image based depth estimation methods \cite{fu2018deep,ming2021deep}, especially in occluded and depth discontinuity regions. 
The success of \dnn in other vision domains \cite{sandler2018mobilenetv2, he2016deep, yu2021bisenet, simonyan2014very} in the last decade has led to the birth of \dnn based stereo methods. Initially researchers worked on optimizing  one or more stereo modules using neural networks \cite{zbontar_stereo_2016, seki_sgm-nets:_2017,park2016look}, but lately end-to-end methods \cite{kendall2017endtoend,chang_pyramid_2018,zhang_ga-net:_2019, xu2022acvnet} have overtaken traditional and single-module based networks.

\begin{figure}[!t]	
	\begin{center}
		\includegraphics[width=\columnwidth]{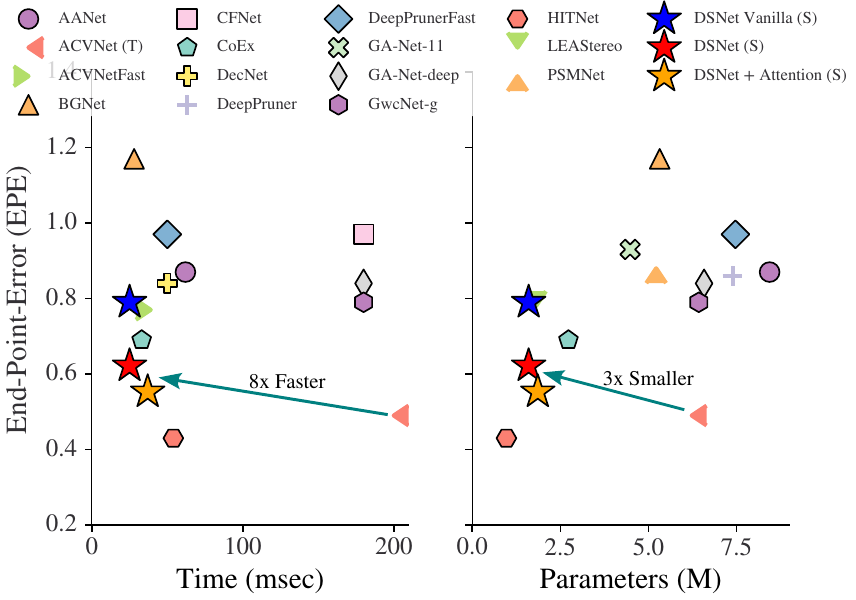}
		\vspace*{-20pt}
	\end{center}		
	\caption{Comparison of our student network with other state of the art methods on \sf test set. Compared to teacher network (\acvnet \cite{xu2022acvnet}), our proposed student network (\dsnet (S)) has $3 \times$ fewer parameters and is $ 8 \times $ faster.}
	\label{fig:comparison}
		\vspace*{-5pt}
\end{figure}

These recent methods can be broadly classified into either \twd or \td methods depending upon the rank of convolution kernels used in later stages for merging information from the views, \ie \twd methods use \twd convolution kernels on a \td fused volume whereas \td ones use \td convolutions on a \fd fused volume.  \td methods \cite{kendall2017endtoend, chang_pyramid_2018, zhang_ga-net:_2019, wang_anytime_2019, duggal_deeppruner:_2019, xu2022acvnet, guo_group-wise_2019} bring the advantage of improved performance in comparison to \twd methods \cite{mayer_large_2016, liang_learning_2018, tonioni2019real, yang_segstereo_2018, song_edgestereo_2018} but at the cost of increased computational budget and bigger model sizes \cite{rahim2021separable, shamsafar2022mobilestereonet}.  This limits their potential for real world applications as they are bulky and slow to run.

Here our goal is to design  stereo matching networks that not only give good performance but are leaner and faster. To this end, we exploit the principle of \kd to distill large \sota \td methods to smaller \td ones.
Although \kd has proven successful in other computer vision domains \cite{liu2020structured, beyer2022knowledge}  for designing real-time lighter student networks, not much progress has been made when it comes to stereo vision \cite{gao2020compact}  -- there exists not a single open source method for distilling stereo networks. One of the main reasons for this lack of progress is that it is not straightforward to define a distillation pipeline in the case of stereo matching because one has to answer many design questions;  ranging from: \textit{(i)} what to choose as student architecture or backbone? \textit{(ii}) from which point(s) in the teacher network to distill the information? \textit{(iii)} how to distill the information from four-dimensional cost volume? \textit{(iv)} what type of loss functions to use for different features representations?, \etc. 

To this end, we answer these questions for building efficient light-weight student networks using \kd.  Overall we make following contributions:
\begin{itemize}
		\itemsep-0.1em 
	\item We  propose a novel \kd pipeline for stereo matching by: \textit{(i)} carefully investigating different configurations for designing efficient  student network; \textit{(ii}) empirically evaluating the importance of using different distillation points; \textit{(iii) } exploiting learned attention weights for distilling cost volume; and \textit{(iv)} by exploring and investigating the influence of  different loss functions on student's network performance. In addition, we design a joint objective function to optimize the student network over different distillation points.  
	\item We do a detailed empirical evaluation of all the design choices for building an efficient student network and \kd.
	\item Consequently, our trained student network (called \dsnet), built using the proposed \kd pipeline, gives reasonable performance compared to state of the art teacher network while having $3 \times $ fewer parameters and being $8\times$ faster. Moreover, compared to performance oriented \sota methods (on \sf dataset) like \psm\cite{chang_pyramid_2018}, \cfnet \cite{shen2021cfnet}, and \leastereo~\cite{cheng2020hierarchical} our student not only gives better performance but is $ 8\times $, $ 5\times $, and $8\times $ faster respectively. On the other hand, compared to speed oriented methods our students perform better than all the tested methods on \kitti and \sf benchmarks while being faster as well. 
\end{itemize}

%

\section{Related Work}

Recently, the computer vision community's focus has shifted from optimizing \dnn for performance to optimizing them for computational requirements \cite{sandler2018mobilenetv2, tan2019efficientnet, yu2021bisenet,hinton2015distilling}, as models are becoming bigger and resource hungry. Given this, numerous efforts have also been made to make stereo networks light-weight and real-time. 

\begin{figure*}[!ht]
	\centering 
	\includegraphics[width=0.9\linewidth]{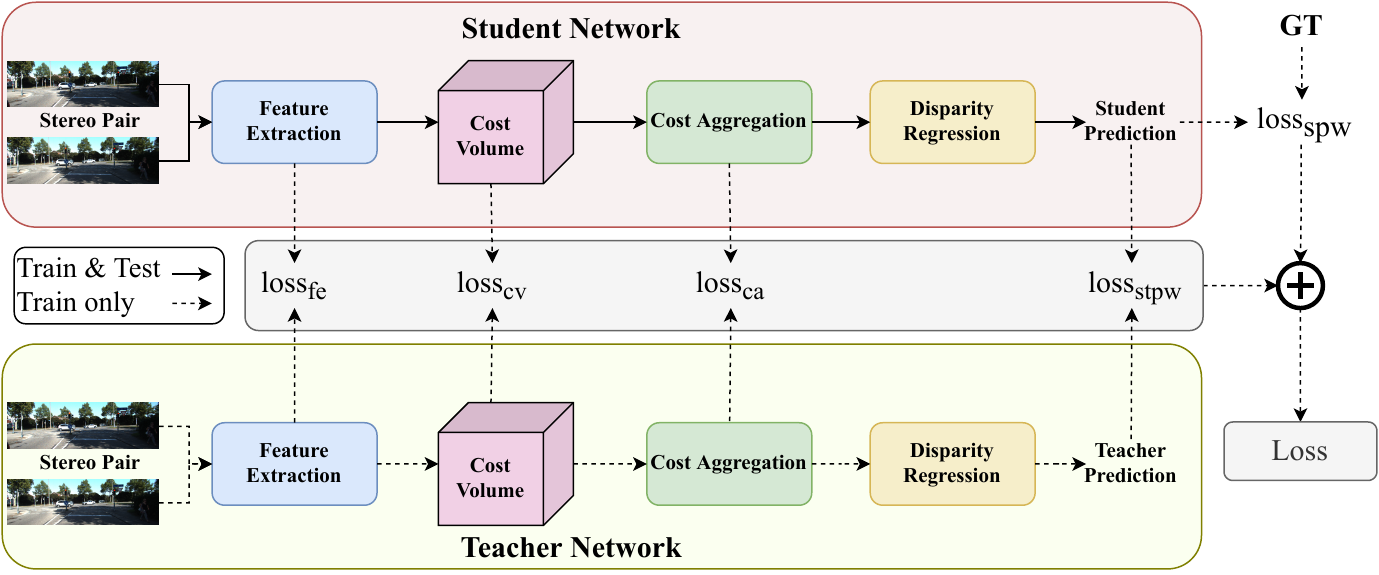}
	\caption{Proposed \kd pipeline for stereo matching. We use different distillation points with different types of loss functions to distill learned information from teacher to student. Here $loss_{fe} $ is feature extraction loss, $loss_{cv}$ is cost volume loss, $loss_{ca} $ is cost aggregation loss, $loss_{stpw} $ is student pixel-wise loss \wrt teacher and $loss_{spw} $ is student pixel-wise loss \wrt ground-truth(GT). $ Loss $ is overall learning objective defined by accumulating each component's loss as explained in \secref{methodology}.}
	\label{fig:architecture}
	\vspace*{-10pt}
\end{figure*}

\gwcnet \cite{guo_group-wise_2019} introduced group-wise correlation to reduce the size of costly 4D volume introduced in \cite{kendall_end--end_2017}, to make the overall network faster and better. Anynet \cite{wang_anytime_2019} reduces the network size by deploying a coarse-to-fine resolution strategy that has been successful in \twd networks. Deeppruner \cite{duggal_deeppruner:_2019} uses patchmatch to narrow down the search range for disparities. MobileStereoNet \cite{shamsafar2022mobilestereonet} and depth-wise separable convolutions \cite{rahim2021separable} have also significantly reduced the stereo network's computational requirements. StereoNet \cite{khamis2018stereonet} uses a very low resolution cost volume to regress disparities at low resolution and then hierarchically improve the disparity maps to full resolution. MADNet \cite{tonioni2019real} introduces an unsupervised real-time stereo network that independently trains the sub-portions of the network.
Although there have been many works to speed up the stereo networks \cite{xu_aanet_2020,yao2021decomposition,tankovich2021hitnet,song2020edgestereo} there is a long way to go, specifically for \td stereo networks which are still very computation demanding. 

Lately, \kd has been very successful in other domains like face recognition \cite{luo2016face}, classification \cite{beyer2022knowledge}, object detection \cite{chen2017learning}, segmentation \cite{liu2020structured}, and semi-supervised \cite{xie2020self} and self-supervised \cite{grill2020bootstrap} learning to train lightweight networks. 
\Kd \cite{hinton2015distilling} is a technique of learning the output and/or intermediate activations of a bigger and stronger network (usually called a teacher network) by a compact and light-weight network called a student network. 
When it comes to stereo matching, very limited research has been done in this direction \cite{gao2020compact}. Compact StereoNet\cite{gao2020compact} uses only a single distillation point without any detail study of impact of different design choices on distilling stereo information. Therefore there is a need for a method that systematically explores different  design choices for designing \KD based stereo methods . Furhtermore, there exists no open source method that can be used as a baseline for reference and further development purposes. In this work, we aim to systematically explore different design choices for designing a real-time student stereo network using the principle of \kd. 

\section{Methodology}
\label{sec:methodology}
To build a \kd pipeline for stereo matching, one needs to answer the following questions:
\begin{itemize}
	\itemsep-0.1em 
	\item What should be the design of the student network?
	\item What network to use as a teacher?
	\item Should we use multiple distillation points or not? If yes, then what should be those distillation points?
	\item What loss functions to use for different types of features having different dimensions?
\end{itemize}
\subsection{Student Network (S)}
We first design a leaner student network by drawing influence from recent state of the art visual recognition and stereo-matching methods \cite{sandler2018mobilenetv2,wang2018pelee,guo_group-wise_2019}.  A typical stereo network generally consists of four modules: feature extraction, cost volume, cost aggregation and disparity regression (\figref{architecture}). The next subsections contain in-depth details on these modules. 

\subsubsection{Feature Extraction:}
The feature extraction module is basically a shared network with \twd convolution operations and is responsible for modeling important features from the input image pair -- \cf \figref{architecture}. We keep our backbone design explicitly simple by restricting the network building blocks set to the one known to be computationally efficient like ReLU, kernel sizes of $3\times3$,~\etc.

Precisely, the complete \twd feature backbone is build from a series of $3\times3$ convolutional layers, $ B1 $ and $ B2 $ blocks  \cite{simonyan2014very,he2016deep, zhang_ga-net:_2019}. The $ B1 $ is constructed from two layers of $3 \times 3 $ convolutions with a simple skip connection. The $ B2 $ block is similar to $B1$ except that, to increase the receptive field size and to reduce the spatial dimensions, we replace the first convolutional layer with a strided convolution and also use a $ 1 \times 1 $ strided convolution in the  skip connection branch.  

To address computational load requirements, we build and test three different variants of \twd feature extraction modules,  namely \bbo, \bbt and \bbd, with $ 21, 18 $ and $ 14 $ convolutional layers, respectively -- note we count the total number of convolutional layers in the module for uniform comparison.  \tabref{student-design} contains the details of \bbo feature extraction module, here Level $ 1 $ and $ 2 $ means spatial resolution of the features maps is being reduced by $1/2$ and $ 1/4 $ with respect to the input height $ (H) $ and width $ (W) $. As a final output of this module, we concatenate features from the last three layers of the network.
\begin{table}[!htb]
	\centering
	\renewcommand{\arraystretch}{1.2}
	\begin{adjustbox}{width=\columnwidth}
		\begin{tabular}{c|c|c|c|c}
			\hline
			{No.} & {Level} & {Input Size} & {Operation}  & {Output Size} \\ \hline
			\hline
			1. & 1 & $ 3 \times H \times W $  & $ 3 \times 3 $ Conv & $ 32 \times H/2 \times W/2 $ \\
			2. & 1 & $ 32 \times H/2 \times W/2 $  & $ 3 \times 3 $ Conv & $ 32 \times H/2 \times W/2 $  \\
			3. & 1 & $ 32 \times H/2 \times W/2 $ & $ 3 \times 3 $ Conv & $ 32 \times H/2 \times W/2 $ \\
		
			4. & 1 & $ 32 \times H/2 \times W/2 $ & B1 & $ 32 \times H/2 \times W/2 $  \\
			5. & 1 & $ 32 \times H/2 \times W/2 $ &  B1 & $ 32 \times H/2 \times W/2 $  \\
			\hline
			6. & 2 & $ 32 \times H/2 \times W/2 $ &  B2 & $ 64 \times H/4 \times W/4 $  \\
			
			7. & 2 & $ 64 \times H/4 \times W/4 $ &  B1 & $ 64 \times H/4 \times W/4 $ \\
			8. & 2 & $ 64 \times H/4 \times W/4 $ & B1 & $ 64 \times H/4 \times W/4 $ \\
			9. & 2 & $ 64 \times H/4 \times W/4 $ & B1 & $ 64 \times H/4 \times W/4 $ \\
			10. & 2 & $ 64 \times H/4 \times W/4 $ & B1 & $ 128 \times H/4 \times W/4 $ \\
			11. & 2 & $ 128 \times H/4 \times W/4 $ & B1 & $ 128 \times H/4 \times W/4 $ \\
			\hline
			Output & 2 & Layer 9, 10, and 11 & Concat & $ 320 \times H/4 \times W/4 $ \\
						\hline
		\end{tabular}
	\end{adjustbox}
	\caption{Architecture of student's (\bbo with 21 layers) feature extraction module.}
	\label{table:student-design}
\end{table}

\subsubsection{Cost Volume Construction:}  This module merges the output feature maps from left and right images to produce four-dimensional output. For cost volume construction we have explored and experimented with two different methods.
In the first method, left and right image feature maps of size $ 320 \times H/4 \times W/4$ are merged via  group-wise correlation to produce an output of $ G \times D/4 \times H/4 \times W/4 $, where $G$ is the correlation group size and $ D $ is the maximum disparity. 

Although this simple method leads to a faster student with good performance however to further improve the performance we have also experimented with another  method. In this method, we integrate learned attention with group-wise correlation to design the cost volume \cite{xu2022acvnet}.  Specifically, first a small attention network is used to learn the attention weights from the output of the feature extraction module. Next these learned weights are then applied to the group-wise correlation volume to refine and improve the overall disparity regression -- \cf Table  \ref{table:ablation-study-64e}. 
\subsubsection{Cost Aggregation:} 
The cost aggregation module takes the output from \fd cost volume and processes it by a series of \td encoder-decoder (ED) networks.   Since this module uses \td convolutions it consumes the majority of the operations in the overall network. To keep the computational foot-print of the student network manageable we also explored different design choices for this module as well. To this end,  we experimented with varying the number of ED networks and the number of \td  layers ($N$) in each ED network. Overall we experimented with six different variants as explained in \secref{design-choices}. \tabref{hg-design} contains the design of a ED network composed of four layers of \td encoder convolutions and two layers of transposed \td decoder convolutions. 
\begin{table}[htb!]
	\centering
		\renewcommand{\arraystretch}{1.3}
	\begin{adjustbox}{width=\columnwidth}
		\begin{tabular}{c|c|c|c|c}
			\hline
			{No.} & {Level} & {Input Size} & {Operation} & {Output Size} \\ \hline
			\hline
			1. & 3 & $N \times D/4 \times H/4 \times W/4 $  & $ 3 \times 3 \times 3 $ Conv & ($N\times 2) \times D/8 \times H/8 \times W/8 $ \\
			2. & 3 & ($N\times 2) \times D/8 \times H/8 \times W/8 $   & $ 3 \times 3\times 3 $ Conv & ($N\times 2) \times D/8 \times H/8 \times W/8 $   \\
			\hline
			3. & 4 & ($N\times 2) \times D/8 \times H/8 \times W/8 $  & $ 3 \times 3 \times 3$ Conv & ($N\times 4) \times D/16 \times H/16 \times W/16 $  \\
			4. & 4 & ($N\times 4) \times D/16 \times H/16 \times W/16 $ & $ 3 \times 3 \times 3$ Conv & ($N\times 4) \times D/16 \times H/16 \times W/16 $  \\
			\hline
			5a. & 3 & ($N\times 4) \times D/16 \times H/16 \times W/16 $ &  $ 3 \times 3 \times 3$ Transpose Conv & ($N\times 2) \times D/8 \times H/8 \times W/8 $  \\
			5b. & 3 & ($N\times 2) \times D/8 \times H/8 \times W/8 $ &  Skip connection from 2. & ($N\times 2) \times D/8 \times H/8 \times W/8 $  \\
			\hline
			6a. & 2 & ($N\times 2) \times D/8 \times H/8 \times W/8 $ &  $ 3 \times 3 \times 3$ Transpose Conv & $N \times D/4 \times H/4 \times W/4 $  \\
			6b. & 2 & $N \times D/4 \times H/4 \times W/4 $ &  Skip connection from 1.  & $N \times D/4 \times H/4 \times W/4 $  \\
			
			\hline
		\end{tabular}
	\end{adjustbox}
	\caption{ \td Encoder-Decoder (ED) based cost aggregation module. Here `$ N $'  is the number of \td convolution filters in a layer.  }
	\label{table:hg-design}
	\vspace*{-15pt}
\end{table}
\subsubsection{Disparity Regression:} In disparity regression, during training, we regress disparity from multiple intermediate network outputs to improve overall features learning inside the network \cite{guo_group-wise_2019}. Precisely, first output of each ED networks is passed through a pair of \td convolutions  and upsampling operations to obtain an output volume of size  $ D \times H \times W $. This volume is then converted to probabilities via \textit{softmax} and passed through \textit{softargmin} to estimate the disparity map of size $ H\times W $. During inference, disparities are estimated only from the output of the final ED network.
\subsection{Teacher Network (T)}  \label{sec:teacher}
As a teacher network, we experimented with different state of the art publicly available stereo matching methods. Specifically, we explored \acvnet \cite{xu2022acvnet}, \gwcnet \cite{guo_group-wise_2019} and \lacnet \cite{liu2022local} as our teacher networks. \acvnet is a current state-of-the-art method\footnote{HITNet \cite{tankovich2021hitnet} is another very strong competing method but has no public implementation available.} and thus is a strong candidate to be considered as a teacher network -- we also find it to be the most promising teacher empirically as well. In addition, we also consider \gwcnet and \lacnet as teachers to evaluate how well the student network can learn coupled with teachers of different strengths and strong ground truth signals. Although we also explored the idea of distilling from multiple teachers, however due to their large memory foot-print we could not fit more than one teacher in the memory with our training setup.
\subsection{Loss Functions} 
For \kd, we investigate many different commonly used loss functions in image-level tasks such as stereo matching, segmentation, mono-depth estimation \etc as well as those used by other distillation networks \cite{liu2020structured, park2019relational, chen2021distilling, passalis2018learning}.   Here we provide details of the loss functions that we have evaluated and which have been finally selected to train our student model. Overall, we find that in \kd for stereo matching not only the type of loss function is important but  also at what point in the network loss function is being used for information distillation is also important -- \cf~\secref{experiments and results}. 

In the following discussion, $ \Phi^S_d $ and $ \Phi^T_d$ represent multi-dimensional student and teacher features respectively and $d$ is feature matrix rank, which  depending on at the distillation point in the network can be of rank 2, 3 or 4. 


\textbf{\smoothl Loss:} 
Recent  stereo methods \cite{gao2020compact,xu2022acvnet} use \smoothl as loss function for learning disparities. \smoothl is  known to be much more robust against outliers and also leads to better learning of the model as it avoids the problem of exploding gradients in some cases~\cite{ren2015faster}.
\begin{equation*}
SmoothL1(\Phi^S_d, \Phi^T_d)=   
\begin{cases}
0.5 \frac{\parallel \Phi^S_d - \Phi^T_d\parallel ^2}{\tau}, & \hspace{-7pt}\text{if } |\Phi^S_d - \Phi^T_d|  < \tau \\
|\Phi^S_d - \Phi^T_d| - 0.5\tau, & \hspace{-7pt}\text{else}
\end{cases}  
\end{equation*}
where we use $\tau=1$. 

\textbf{\logl Loss:}
One of the downsides of using \ltwo or \lone (or both) as loss function for learning disparities is that these error functions either have constant influence on the learning (in case of L1) or proportional to error (in case of L2) without considering whether the error computed involves outliers or not. To this end,  following the works in mono-disparity estimation \cite{Hu2018,watson2019,Jae-Han2018single}, we use $\log$ as a squeezing function on top of \lone loss to penalize small errors severely as well. Precisely,
\begin{equation*}
LogL1(\Phi^S_d, \Phi^T_d) =  \log ( |\Phi^S_d - \Phi^T_d| + \epsilon)
\end{equation*}
To keep loss values positive,  we set $\epsilon \ge 1$. Our experiments show that \logl loss leads to better performance and is more stable during the training.

\textbf{\cosl Loss:}
For matching features in the earlier layers and cost volumes, we use a cosine similarity based loss function, where
\begin{equation*}
Cosine Loss(\Phi^S_d, \Phi^T_d)=    1 - \cos( \Phi^S_d , \Phi^T_d)
\end{equation*}

\textbf{KL Divergence (\kld) Loss:} 
In this loss function, we consider the teacher predicted disparities (or features) as true class probabilities ($p_T$) and student disparities as predictions probabilities ($p_S$) and measure the Kullback-Lieber divergence (\kld) as a loss.  \Ie
\begin{equation*}
KLDLoss(S\parallel T) = p_T\log p_T - p_T\log p_S\label{eq:kld} 
\end{equation*}

In addition, we also evaluated many other commonly used loss functions for \kd, however, in our experiments these losses produced inferior performance so they are not reported any further.  For instance, we also tested pair-wise loss \cite{liu2020structured}, Relational Knowledge Distillation (RKD) \cite{park2019relational}, Hierarchical Context Loss (HCL) \cite{chen2021distilling} and Probabilistic Knowledge Transfer(PKT) \cite{passalis2018learning} losses. 
%
\subsection{Distillation Points} 
Initially in our baseline student network we used only one point to distill the output  of the teacher's network -- here distilling the output means supervising the student to match the teacher's output via a loss function. However the performance of the student network was too weak, thus to improve its performance we started exploring different distillation points and different loss functions (due to different type and dimensionality of the features) for each of the distillation points -- \cf \figref{architecture}.

For distilling the information from the feature extraction module, we investigated two different distillation points, one from the early layers (from layer 3 and 5 -- \cf \tabref{student-design}) of network and  one from the later layers (layers 9, 10 and 11) of network.  Since cost volume is one of the most important modules in the overall pipeline, as it sets the base for learning of the disparities in the following modules, we set our second distillation point at the output of cost volume module.  The third distillation point was set at the output of the cost aggregation module. Finally the last distillation point was set at the output of the disparity regression module to match the student predicted disparities with the teacher's one. All of these choices were thoroughly validated via detailed ablation study -- \cf \secref{design-choices}.

\subsection{Learning Objective Function}
Finally, based on all the distillation points and teacher and student network outputs, we define the overall objective function as:
\begin{equation*}
\ell(S,T,GT)= 
\begin{cases}
\lambda \ell_{fe}(S,T) + \lambda \ell_{cv}(S,T) + \\
\lambda_{ca} \ell_{ca}(S,T) + \lambda_{spw} \ell_{spw}(S,GT) + \\
\lambda_{stpw} \ell_{stpw}(S,T)
\end{cases}
\end{equation*}
 Here $ S, T$ and $ GT $ represent student, teacher and ground truth, respectively; and $\ell_{fe} $, $ \ell_{cv} $, $ \ell_{ca} $ are distillation point losses from the feature extraction, cost volume,  and cost aggregation modules, respectively.  $ \ell_{spw}$ represents the student's pixel-wise loss \wrt $ GT $ while $ \ell_{stpw} $ represents the student's pixel-wise loss \wrt $ T $.   We set $ \lambda = 0.1 $, $ \lambda_{ca} = 0.1 $, $ \lambda_{spw} = 0.4 $ and $ \lambda_{stpw} = 0.4 $ for our final student network.


\section{Experiments and Results}
\label{sec:experiments and results}
In this section, we first introduce the datasets used for the training and evaluation of our experiments. We then present the implementation details, followed by the discussion on different design choices and results.
\subsection{Datasets}
For the training and evaluation of our models we use two well-known stereo benchmarks datasets, namely \sf \cite{mayer_large_2016} and \kittif ~\cite{menze_object_2015}. For measuring the generalization capability of our setup and trained student models we use two more stereo datasets, namely \ethz \cite{schoeps2017cvpr} and \mb \cite{scharstein2014high}.

\textbf{\kitti} \cite{geiger2012we, menze_object_2015} is a real-world dataset of driving scenes with sparse annotations. \kitti 2012 contains 194 training and 195 test pairs while \kittif contains $200$ image pairs for training and testing each, with resolution $376 \times 1240$ pixels. 

\textbf{\sf}\cite{mayer_large_2016} is a large scale synthetic dataset with dense annotations. It contains 35,454 training images and 4,370 test images of size $540 \times 960$.

\textbf{Evaluation metrics:}  We report mutliple evaluation metrics including D1, End-Point-Errors  (EPE)~\cite{mayer_large_2016,menze_object_2015} and  K-\textit{px} errors.  EPE (reported in \textit{px} units) is the average disparity error. D1 is the percentage of pixels where the \dpix error is greater than  $ 5 \% $ of the ground-truth. K-
\textit{px} is the percentage of pixels where the error is greater than $K$ pixels, for our experiments $K \in \{1,2,3,4\}$. We also report the number of parameters in millions  (M) and number of operations/\macs  (Multiply-ACcumulate operations) in Giga-MACs  (GMACs) for comparison\footnote{ $ 1  \mbox{MAC} =  1 $ multiplication + $ 1 $ addition. These operations are measured using https://github.com/sovrasov/flops-counter.pytorch}. For all of the evaluation metrics, lower values indicates better results.

\subsection{Implementation Details}\label{sec:implementation_details}
For network training, we use the Adam optimizer  ($ \lambda=10^{-4} $, $ \beta_1=0.9  $ and $ \beta_2=0.999 $). All the training images are cropped to the size of $256 \times 512$ with the maximum disparity set to $192$. Furthermore, all the images are standardized using the mean and standard deviation of the ImageNet dataset \cite{deng2009imagenet}. Our final model is trained with a batch size of $ 8 $ on $ 4 $ GPUs. For the \sf dataset, we train all our models for  $ 64 $ epochs (like \cite{xu2022acvnet}) with an initial learning rate of $ 10^{-4} $ that is reduced to half at the $ 20,32,40,48$ and $56^{th}  $ epochs. 
For the \kittif dataset, we fine-tuned our pre-trained \sf model for $ 500 $ epochs. The initial learning rate was set to $10^{-4} $ which was later decreased to $ {1}/{5^{th}} $ after $ 250 $ epochs. 

\begin{table}[htb!]
	\centering
	\begin{adjustbox}{width=\columnwidth}
		\npdecimalsign{.}
		\nprounddigits{2}
		\begin{tabular}{l|c|c|n{2}{3} }
			\hline
			Variant & {Params ($M$)} & {MACs ($G$)} & {EPE ($px$)}  \\ \hline
			\bmark-ED$_3$-N$_{32}$ & 6.52 & 246.27 & 0.73 \\
			\hline
			\bbo & 4.43 & 211.12 & 0.7567 \\
		\bbt   & 4.26 & 207.48 & 0.7733 \\
		\bbd & 4.12 & 200.84 & 0.7874 \\
		\hline

			\hline
			\bbo-ED$_3 $  & 4.43 & 211.12 & 0.7567 \\
			\bbo-ED$_2$ & 3.29 &  165.50 & 0.7734 \\
			\bbo-ED$_1$ & 2.16 & 120.06 & 0.83 \\
			\hline
			\bbo-ED$_2$-N$_{24}$ & 2.25 & 108.7 &   0.78\\
			\rowcolor{blue!12}
			\bbo-ED$_2$-N$_{16}$ & 1.50 & 67.20 &   0.79\\
			\bbo-ED$_2$-N$_8$ & 1.05 &  41.15 &  0.85 \\

			\hline
		\end{tabular}
	\end{adjustbox}
	\caption{Comparison of different student variants performance on \sf dataset. For all the metrics, lower means better.}
	\label{table:student-variants}
\end{table}
\subsection{Evaluation of Design Choices}\label{sec:design-choices}
\textbf{Student Network (S):} 
As discussed in \secref{methodology}, to reduce the computational complexity, we designed and tested student networks with different numbers of layers in the feature extraction module and different numbers of ED networks in the cost aggregation module.  Initially, we started with a very strong baseline student network (\bmark-ED$_3$-N${32}$) containing 56 convolutional layers in the feature extraction module and 3 ED networks (with each having 32 \td convolutional layers) in the cost aggregation module -- this network was trained using \smoothl loss. This network was then iteratively tailored down to have reasonable performance while maintaining a small computational footprint.

We started by reducing the number of layers in \td feature extraction backbone. To this end, we trained three different variants \bbo, \bbt, \bbd with 21, 18 and 14 numbers of convolution layers in the backbone. As results in \tabref{student-variants} show, all of the variants give  comparable performance with a very small difference in parameters and operations. Therefore, we choose to proceed with \textit{\bbo} as having more layers can be helpful when we further reduce the student size in other parts of the network.  

As observed by us and many others \cite{rahim2021separable,shamsafar2022mobilestereonet}, the majority of operations in a \td stereo network are consumed by the ED networks in the cost aggregation module. So after finalizing the number of layers in the backbone network, next we explored reducing the number of ED networks in cost aggregation from 3, to 2 and eventually to 1.  \tabref{student-variants} shows the results of different variants of these student networks.  As can be observed from the results, decreasing the number of ED networks leads to loss in performance however the drop in performance going from \textit{ED$_3$} to \textit{ED$_2$} is not as significant relative to the reduction in the number of \macs. However going from \textit{ED$_2$} to \textit{ED$_1$} has noticeable impact on the performance, as a result we use \textit{ED$_2$} as a cost-aggregation module for our candidate student network(s). 

After finalizing the number of layers in the backbone network and number of ED networks as a next step we explored reducing the number of $\td  $ convolutional filters.  Precisely, starting from a base student network \textit{\bbo-ED$_2$-N$_{32}$} we further reduce the student computational footprint by lowering the number of $\td$ filters from 24 to 16 and finally to 8. As it can be noted, reducing the number of $\td$ filters lead to significant drop in \macs but with noticeable performance drop, from 0.78 \epe to 0.85.  Consequently, we finally chose \textit{\bbo-ED$_2$-N$_{16}$} as our candidate student network (\dsnet Vanilla) for the best compromise between performance and speed.

\begin{table*}[htb!]
	\centering
	\begin{adjustbox}{width=0.8\textwidth}
		\npdecimalsign{.}
		\nprounddigits{2}
		\begin{tabular}{l|c|c|c|c|c|c|n{2}{3}|n{2}{3}|n{2}{3}|n{2}{3}|n{2}{3}}
			\hline
			\makecell{Loss Function \\ for added module} & {\spw} & \makecell{\cv} & \makecell{\fe} & {\stpw}  & \makecell{\ca} & \makecell{\cv+\\Attention} & {EPE ($px$)} & {D1 ($\%$)}  & {\dpix ($\%$)}  & {\tpix ($\%$)} & {\opix ($\%$)}\\ \hline
			\hline
			
			\logl & \checkmark &  &   &  &  &    & 0.7558 & 2.602 & 3.086 & 3.889 & 6.049  \\
			
			\cosl & \checkmark & \checkmark  &   &  &  &   & 0.7135 & 2.488 & 2.975 & 3.796 & 6.038 \\			
			
			\cosl & \checkmark & \checkmark  & \checkmark &  &  &  & 0.7124 & 2.477 & 2.967 & 3.783 & 6.009 \\		

			\smoothl & \checkmark & \checkmark  & \checkmark & \checkmark &  &   & 0.6849 & 2.445 & 2.969 & 3.831 & 6.205   \\	
			
			\kld & \checkmark & \checkmark  & \checkmark & \checkmark & \checkmark &  & 0.6216 & 2.204 & 2.7 & 3.539 & 5.9   \\
			
			\cosl & \checkmark & \checkmark  & \checkmark & \checkmark & \checkmark & \checkmark  & 0.5529 & 1.99 & 2.434 & 3.175 & 5.17   \\
				
			\hline
		\end{tabular}
	\end{adjustbox}

	\caption{Impact of using different loss functions and distillation points on the performance of \dsnet.}
	\label{table:ablation-study-64e}
	\vspace*{-1em}
\end{table*}

\textbf{Distillation Points and Losses:}
Although our chosen candidate (\dsnet Vanilla)  has a lower computational footprint, it has weaker performance compared to the teacher and other state of the art \td stereo methods.  For instance, just trimming down the student network from the baseline results in the rise of \epe from 0.73 to 0.79. We recover this lost performance by introducing different distillation points and using a well-tuned loss function for each distillation point.  \tabref{ablation-study-64e} shows the final chosen distillation points and the corresponding loss functions used in \dsnet. It is important to mention that all these options were thoroughly investigated via detailed experimentation. 

Overall, it is important to use  multiple distillation points along with a matching loss function to achieve the best performance. For distilling the feature information from teacher to student (in network's inner distillation points) \cosl loss is the best choice for the loss function followed by \kld loss. Distilling backbone features only from earlier layers gave better performance relative to later layers features. For matching disparities between student and ground-truth (\spw) \logl is the best choice while using  \smoothl to measure pixel-wise loss (\stpw) between predicted disparity maps of student and teacher networks gives better performance. 
%
%

Although adding the attention learning layers in the student (\dsnetpa) and distilling the attention weights from the teacher network leads to noticeable improvement in the performance. However, this comes at the cost of increased computational footprint. For instance, the inference time of \dsnetpa increases to 37\ms relative to 25\ms of \dsnet. 

\textbf{Teacher Network (T):}
As discussed in \secref{teacher}, we experimented with different models as teacher networks. Overall, we observe that having a strong teacher  \acvnet \cite{xu2022acvnet}, as expected, results in better learning of the student network -- substituting \acvnet with  \gwcnet \cite{guo_group-wise_2019}) as teacher network increases the \epe from 0.62 to 0.74. We also note that the design of cost volume also plays an important role, for instance \gwcnet turns out to be a better teacher than \lacnet, although \lacnet is a better performing method than \gwcnet (replacing \gwcnet with \lacnet increases the \epe to 0.79).  From this, we can conclude that the cost volume plays a vital role in the overall stereo pipeline and aligning the design between the cost volumes of student and teacher is important for distillation.

\textbf{Impact of Longer Training Regime:} Although training the student for longer epochs (up to 300 epochs on \sf dataset like  \cite{tankovich2021hitnet} ) leads to student matching the teacher performance. However, we observed that this leads to overfitting and poor generalization capabilities across the datasets. Moreover, training for these many epochs requires three weeks on our setup so this was not explored any further.

\subsection{Results}
This section presents the quantitative and qualitative results of the final student model(s) on different datasets. Following \cite{tankovich2021hitnet,bangunharcana2021correlate,xu2022acvnet} we perform comparisons at two levels, \ie (i) with the performance oriented methods (having runtime $>$ 100 \ms); and (ii) with speed oriented \td stereo methods. 

\textbf{\sf:} 
\tabref{sf-quantity-speed} (top-half) compares the results of our trained \dsnetpa with teacher and performance oriented \sota stereo matching methods on \sf test set. We can observe that despite having $2 \text{ to } 3 \times$ fewer \macs and parameters our method performs better than state of the art methods like \leastereo, \cfnet, \gwcnet, \etc. Moreover, this student network gives comparable performance compared to the teacher network \acvnet while having $3\times$ fewer parameters and operations. With respect to model size, our network also has the smallest memory footprint (7MB) compared to all the other methods tested.
\begin{table}[htb!]
		\centering		
		\begin{adjustbox}{width=\columnwidth}
		\npdecimalsign{.}
		\nprounddigits{2}
		\begin{tabular}{l|ccc|rr}		
			\hline
			{Method} & {EPE($px$)} &  \makecell{MACs\\($G$)} & \makecell{Params\\($M$)} & \makecell{More \\ MACs}  & \makecell{More \\ Params}\\
			\hline
			\hline
			GCNet\cite{kendall_end--end_2017} & 1.84  & 718.01 & 3.18    & 8$\times$ & 2$\times$ \\
			PSMNet\cite{chang_pyramid_2018}	& 0.88 & 256.66 & 5.22    & 3$\times$ & 3$\times$ \\
			GA-Net-deep\cite{zhang_ga-net:_2019} & 0.84 & 670.25 & 6.58    & 7$\times$ & 4$\times$ \\
			GA-Net-11\cite{zhang_ga-net:_2019} & 0.93  & 383.42 & 4.48    & 4$\times$ & 2$\times$\\
			GwcNet-g\cite{guo_group-wise_2019} & 0.79 & 246.27 & 6.43    & 3$\times$ & 3$\times$ \\
			DeepPruner\cite{duggal_deeppruner:_2019} & 0.86  & 129.23 & 7.39& 1$\times$ & 4$\times$ \\ 
			CFNet\cite{shen2021cfnet} & 0.97  &  177.50 & 23.05& 2$\times$ & 12$\times$ \\ 
			LEAStereo\cite{cheng2020hierarchical} & 0.78  & 156.5 & 1.81 & 2$\times$ & 1$ \times $ \\ 
			HITNet$^{\dagger}$\cite{tankovich2021hitnet} & \textbf{0.43} &   146.01 & \textbf{0.97} & 2$\times$ & 1$\times$ \\ 
			ACVNet (T)\cite{xu2022acvnet} & 0.49   & 240.05 & 6.23& 3$\times$ & 3$\times$ \\ \hline
			\rowcolor{blue!12}
			\dsnetpa (S) & 0.55 & \textbf{91.33} & 1.86  & - & -\\ \hline 			

			\hline
		
			DeepPrunerFast\cite{duggal_deeppruner:_2019} & 0.97 & 51.83 & 7.47 & 0.8 & 4.7 \\
			AANet\cite{xu_aanet_2020} & 0.87 & 188.76 & 8.44 & 2.8 & 5.3 \\
			BGNet\cite{xu2021bilateral} & 1.17 & \textbf{20.33 }& 5.32 & 0.3 & 3.3 \\
			DecNet$^\dagger$ \cite{yao2021decomposition} & 0.84 & - & - & - & - \\
			CoEx\cite{bangunharcana2021correlate} & 0.69 &{61} & 2.73 & 0.9 & 1.7 \\
			Fast-ACVNet$^\dagger$ \cite{xu2022acvnet} & 0.64 & - & - & - & - \\ \hline
			
			\rowcolor{blue!12}
			\dsnet  (S)& \textbf{0.62}	& 67.20 & \textbf{1.60}  & - & -\\ \hline
			
		\end{tabular}
	\end{adjustbox}
\caption{Comparison of our student (\dsnet (S)) with performance and speed oriented \sota  stereo methods on SceneFlow test set. \macs and Params calculations of all these methods have been reproduced using identical size input images. $^\dagger$without public implementation. }
	\label{table:sf-quantity-speed}	
\end{table}

\begin{figure}[t]
	\centering
	\begin{adjustbox}{width=\columnwidth}
\begin{tabular}{m{1em}ccm{1em}}
	\rotatebox[origin=c]{90}{\textit{Left}}	& \makecell{\includegraphics[width=0.48\columnwidth]{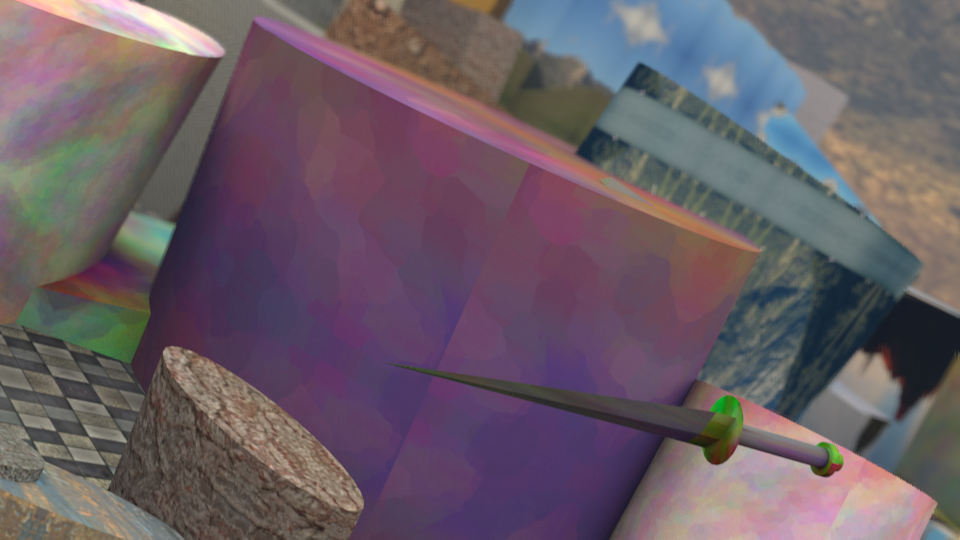}} &		  \makecell{\includegraphics[width=0.50\columnwidth]{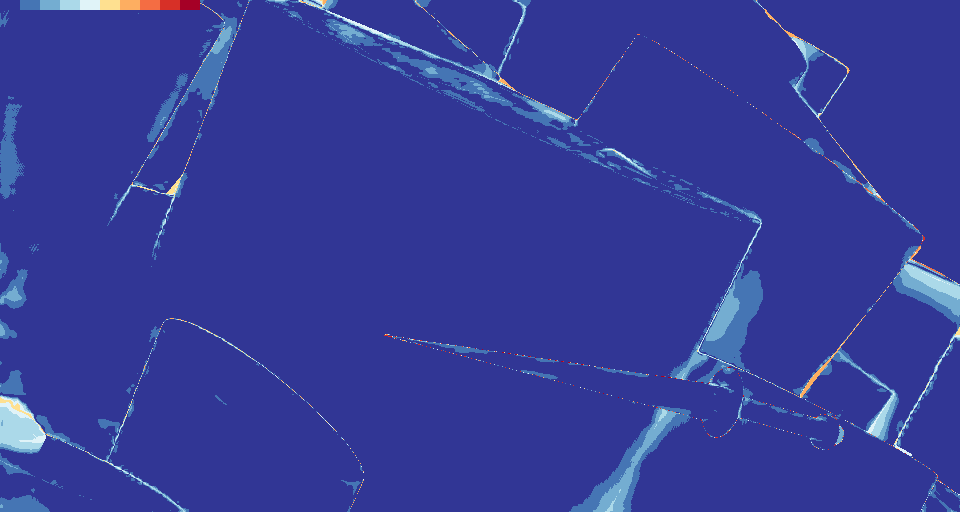}} &  
 \rotatebox{90}{\textit{Error Map}}	
	\\
	\rotatebox{90}{\textit{GT}}	&		\makecell{\includegraphics[width=0.48\linewidth]{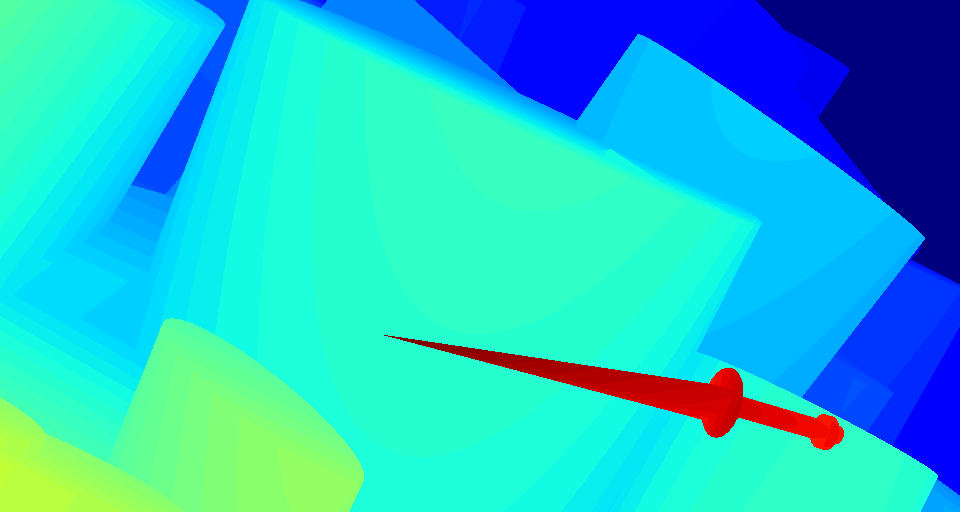}} &		\makecell{\includegraphics[width=0.50\linewidth]{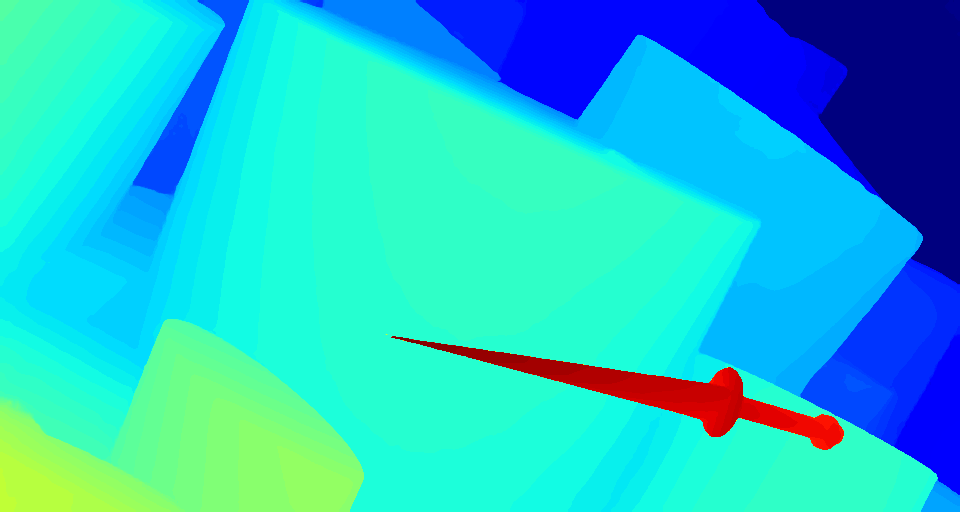}}
	& 
  \rotatebox{90}{\textit{Prediction}}	
	\\
	\rotatebox{90}{\textit{Left}}	&			\makecell{\includegraphics[width=0.48\columnwidth]{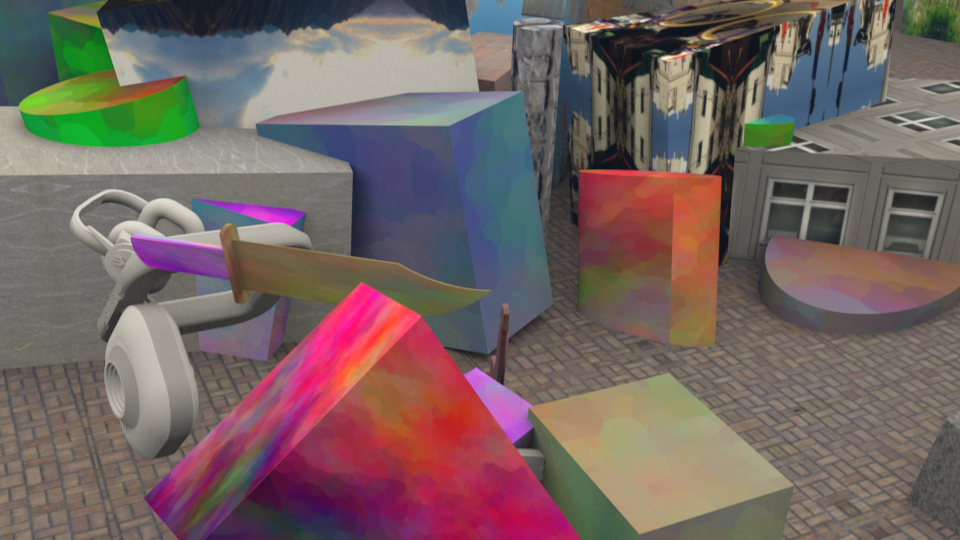}} &		\makecell{\includegraphics[width=0.50\columnwidth]{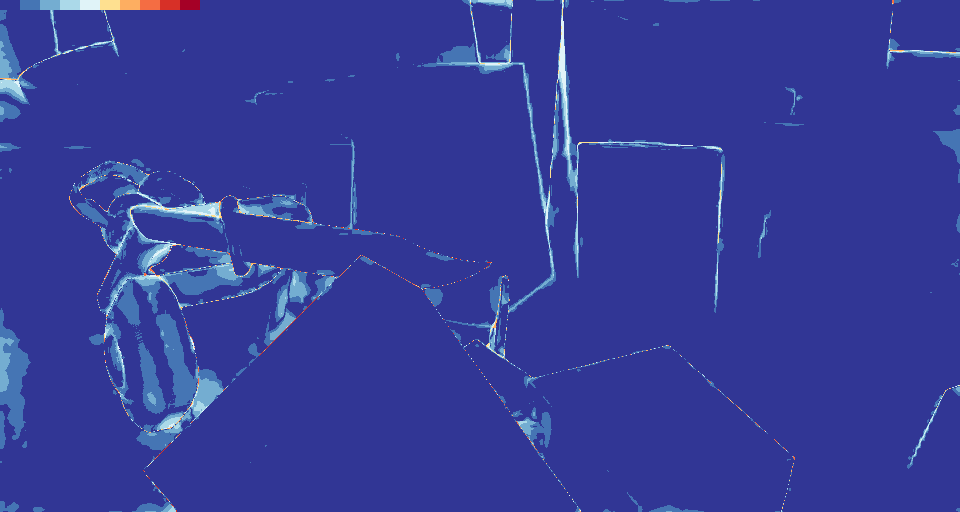}} 
	&  
	\rotatebox{90}{\textit{Error Map}}	
	\\
	\rotatebox{90}{\textit{GT}}	&			\makecell{\includegraphics[width=0.48\linewidth]{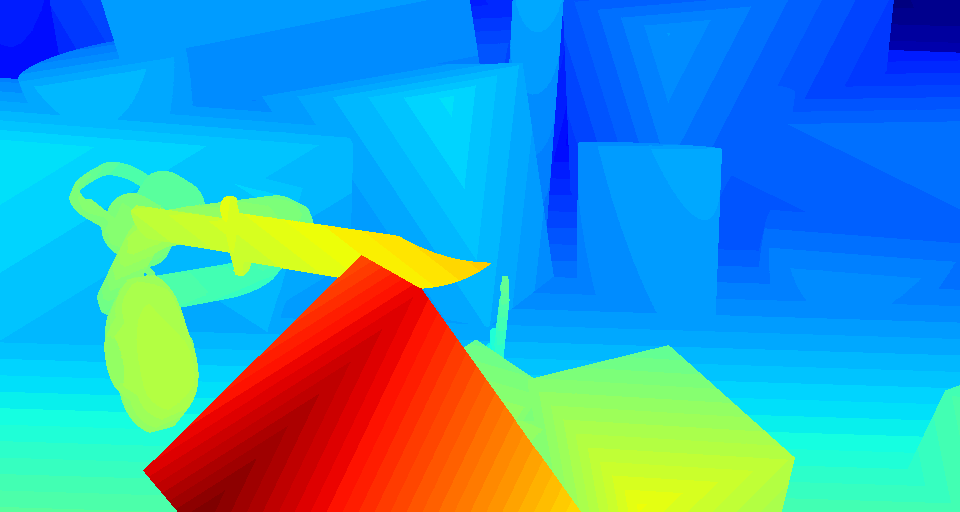}} &		\makecell{\includegraphics[width=0.50\linewidth]{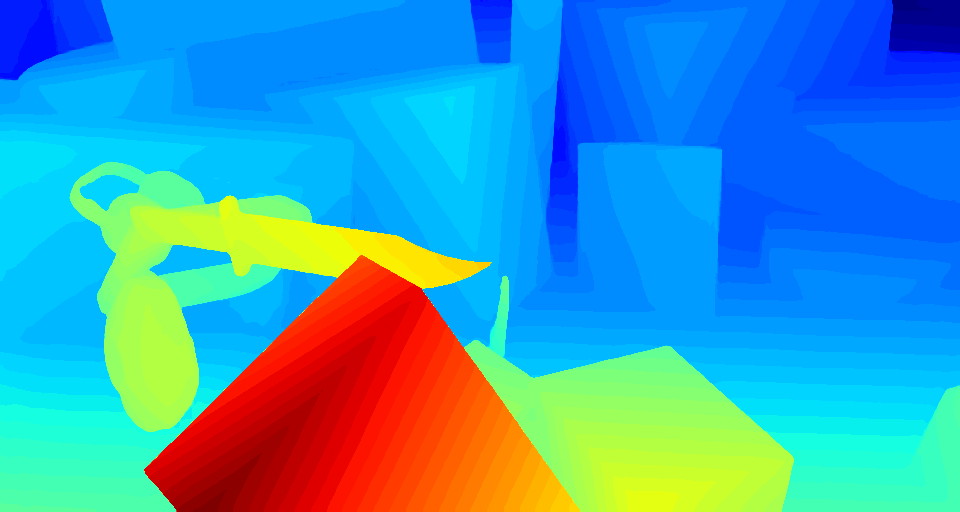}}
	&  
	\rotatebox{90}{\textit{Prediction}}	
	\\
\end{tabular}
\end{adjustbox}
\caption{Qualitative results on sample SceneFlow test images. In error maps, darker red means high disparity errors and as the color gets more blue (darker blue) it represents lower disparity errors. }
\label{fig:sf-quality}
\vspace*{-10pt}
\end{figure}

\tabref{sf-quantity-speed} (bottom-half) compares \dsnet with other speed oriented methods. Here \dsnet outperforms all the existing methods while being faster. This validates that using our proposed distillation pipeline one can train a faster and leaner network with better performance.

\figref{sf-quality} shows qualitative results of our method. From the disparity errors we can observe that our method gives reliable disparity estimates.

\textbf{\kitti:}
\tabref{kitti15-benchmark} compares the result of our submitted method with other methods on the benchmark datasets. For \kittif, We report the D1 error for foreground \textit{(fg)}, background \textit{(bg)} and all pixels from the benchmark page.   For KITTI2012, we report \fpix and \dpix (noc and all) errors.
We can draw similar conclusions as in the case of \sf dataset. Compared to performance oriented methods, \dsnetpa is significantly faster (\eg $6 \times$ relative to \acvnet) with reasonable performance -- in fact it gives identical performance to CFNet and GANet while about $5 \times$ faster. Compared to speed oriented methods, \dsnet has consistently lower  errors  on both the datasets compared to all the other real-time stereo methods including Fast-\acvnet\cite{xu2022acvnet},  AANet\cite{xu_aanet_2020} and DeepPrunerFast\cite{duggal_deeppruner:_2019} while being faster.

\begin{table}[htb!]
	\centering
	\begin{adjustbox}{width=\columnwidth}
		\begin{tabular}{c|c|cccc|ccc|c}
			\hline
			\multirow{3}{*}{Target}&\multirow{3}{*}{Method}&\multicolumn{4}{c|}{KITTI 2012 \cite{geiger2012we}}&\multicolumn{3}{c|}{ KITTI 2015 \cite{menze_object_2015}}&\\
			\cline{3-9}
			&&\dpix (noc)&\dpix(all)&\fpix(noc)&\fpix(all)&D1-bg&D1-fg&D1-all&\makecell{Runtime \\  (ms)} \\
			\hline
			
			\multirow{7}{*}{ \rotatebox{90}{\textit{Performance}}}&{GCNet} \cite{kendall_end--end_2017}&1.77&2.30&1.36&1.77&2.21&6.16&2.87&900 \\
			&{PSMNet \cite{chang_pyramid_2018}}&1.49&1.89&1.12&1.42&1.86&4.62&2.32&$310^{\ddagger}$\\
			&{GwcNet~\cite{guo_group-wise_2019}}&1.32&1.70&0.99&1.27&1.74&3.93&2.11&$180^{\ddagger}$\\ 
			&{GANet-deep~\cite{zhang_ga-net:_2019}}&1.19&1.60&0.91&1.23&1.48&3.46&1.81&180\\ 
			&EdgeStereo\cite{song2020edgestereo}&1.46&1.83&1.07&1.34&1.84&3.30&2.08&320 \\
			&CFNet~\cite{shen2021cfnet}&1.23&1.58&0.92&1.18&1.54&3.56&1.88&180 \\
			&LEAStereo~\cite{cheng2020hierarchical}&\textbf{1.13}&\textbf{1.45}&\textbf{0.83}&\textbf{1.08}&1.40&\textbf{2.91}&\textbf{1.65}&300\\
			&ACVNet (T) \cite{xu2022acvnet}&\textbf{1.13}&1.47&0.86&1.12&\textbf{1.37}&3.07&\textbf{1.65}&200$^{\ddagger}$\\
			
			\hline
			\multirow{10}{*}{ \rotatebox{90}{\textit{Speed}}}&{DispNetC \cite{mayer_large_2016}}&4.11&4.65&2.77&3.20&4.32&4.41&4.34&60\\
			&{DeepPrunerFast\cite{duggal_deeppruner:_2019}}&-&-&-&-&2.32&3.91&2.59&$50^{\ddagger}$\\
			&{AANet\cite{xu2022acvnet}}&1.91&2.42&1.46&1.87&1.99&5.39&2.55&62\\
			&{DecNet~\cite{yao2021decomposition}}&-&-&-&-&2.07&3.87&2.37&50\\
			&{BGNet~\cite{xu2021bilateral}}&1.77&2.15&-&-&2.07&4.74&2.51&$28^{\ddagger}$\\
			&{BGNet+~\cite{xu2021bilateral}}&1.62&2.03&1.16&1.48&1.81&4.09&2.19&$35^{\ddagger}$\\
			&{CoEx~\cite{bangunharcana2021correlate}}&1.55&1.93&1.15&1.42&1.79&3.82&2.13&$33^{\ddagger}$\\
			&{HITNet~\cite{tankovich2021hitnet}}&1.41&1.89&1.14&1.53&1.74&\textbf{3.20}&1.98&$54^{\ddagger}$\\
			
			&{Fast-ACVNet }\cite{xu2022acvnet}&1.68&2.13&1.23&1.56&1.82&3.93&2.17&39$^{\ddagger}$\\
			\rowcolor{blue!12}
			
			&\dsnet (S)&1.36&1.79&1.03&1.37&1.70&3.56&2.01&\textbf{25}$^{\dagger}$\\
			\rowcolor{blue!12}
			&\dsnetpa (S)&\textbf{1.22}&\textbf{1.63}&\textbf{0.93}&\textbf{1.24}&\textbf{1.56}&3.84&\textbf{1.94}&\textbf{37}$^{\dagger}$\\
			\hline
		\end{tabular}
	\end{adjustbox}
	\caption{ Quantitative comparison of our students (S) with performance and speed oriented \sota  stereo methods on KITTI2012 and KITTI2015 benchmark datasets. 
		$^\ddagger$denotes the runtime on identical hardware with same image sizes. $^\dagger=$ TensorRT optimized models.}
	\label{table:kitti15-benchmark}
	
\end{table}

\begin{table}[!htb]
	\centering
	\resizebox{\columnwidth}{!}{
		\begin{tabular}{c|c|c|c|c}
			\hline
			Method    & \begin{tabular}[c]{@{}c@{}}KITTI2012\\ D1-all \end{tabular} & \begin{tabular}[c]{@{}c@{}}KITTI2015\\ D1-all \end{tabular} & \begin{tabular}[c]{@{}c@{}}Middlebury\\ \tpix (bad 2.0) \end{tabular} & \begin{tabular}[c]{@{}c@{}}ETH3D\\  \opix (bad 1.0) \end{tabular} \\ \hline
			PSMNet \cite{chang_pyramid_2018}   & 15.1  & 16.3  & 39.5   & 23.8  \\ 
			GWCNet \cite{guo_group-wise_2019}    & 12.0  & 12.2  & 37.4   & 11.0  \\ 
			CasStereo \cite{gu2019cas} & 11.8  & 11.9  & 40.6   & \textbf{7.8}   \\ 
			GANet \cite{zhang_ga-net:_2019}    & 10.1  & 11.7  & 32.2   & 14.1  \\ 
			DeepPrunerFast~\cite{duggal_deeppruner:_2019} & 16.8  & 15.9 & 30.8 & -\\
			BGNet~\cite{xu2021bilateral} & 24.8  & 20.1 & 37.0 & -\\
			CoEx~\cite{bangunharcana2021correlate} & 13.5  & 11.6 & 25.5 & -\\
			Fast-ACVNet\cite{xu2022acvnet} & 12.4  & 10.6 & 20.1 & -\\ \hline
			\rowcolor{blue!12}
			\dsnet (S) & \textbf{8.6}  & \textbf{8.8} & \textbf{14.24} & 9.2\\ \hline
			
		\end{tabular}
	}
	\caption{Generalization performance of student network on cross-domain datasets. Here the network is only trained on \sf training dataset and evaluated on these diverse real worlds datasets.}
	\label{table:generalization}
\end{table}

%
\textbf{Generalization Capacity of the Student Network}
	\label{sec:generalization}
	To quantify cross-data generalization performance of the \dsnet, we train it on the synthetic \sf dataset and evaluate it on three different datasets.  \tabref{generalization} shows that the \dsnet generalizes extremely well and outperform all the tested methods as well. Thus using the proposed \kd pipeline one can train a student network that not only performs well on the trained dataset but can also generalize well across the datasets.

\section{Conclusions}
In this work we have presented a novel knowledge distillation pipeline by combining insights from stereo methods with general knowledge-distillation techniques. To this end, we have done a thorough systematic study of different design choices for building a leaner and faster stereo network with good performance. We find that distilling stereo networks require careful design of backbone along with right selection of distillation points and loss functions. Our trained student networks compete with performance oriented methods while being significantly faster. Moreover, when compared to speed oriented methods they outperform all of them. As the presented pipeline is generic it can be easily used as a baseline for trimming down real world \td stereo networks. Since no public existing work is already available on this topic we believe this work can serve as a baseline to build further strong knowledge distillation methods for the stereo networks.

\bibliographystyle{IEEEtran}
\bibliography{IEEEabrv,Stereo}

\end{document}